\documentclass[a4paper,11pt]{article}

\usepackage{graphicx}  
\usepackage{url}       
\usepackage{times}
\usepackage{amsmath, amsfonts, amssymb}
\usepackage{braket}
\usepackage{tikz}
\usepackage{qtree}
\usepackage{wrapfig}
\usepackage{fixltx2e}
\usepackage{color}
\usepackage[margin=25mm]{geometry}

\usetikzlibrary{positioning}

\title{Harmonic Grammar in a DisCo Model of Meaning}
\date{}

\author{Martha Lewis, Bob Coecke \\
       Department of Computer Science, University of Oxford \\
\texttt{<firstname>.<lastname>@cs.ox.ac.uk}	}

\begin{document}
\maketitle
\thispagestyle{empty}
\pagestyle{empty}

\section{Introduction}
The model of cognition  developed in \cite{smolensky2006} seeks to unify two levels of description of the cognitive process: the connectionist and the symbolic. The theory developed brings together these two levels into the Integrated Connectionist/Symbolic Cognitive architecture (ICS). \cite{clark2007} draw a parallel with semantics where meaning may be modelled on both distributional and symbolic levels, developed by \cite{coecke2010} into the Distributional Compositional (DisCo) model of meaning. In the current work, we revisit Smolensky and Legendre (S\&L)'s model. We describe the DisCo framework, summarise the key ideas in S\&L's architecture, and describe how their description of \emph{harmony} as a graded measure of grammaticality may be applied in the DisCo model.

\section{Distributional Compositional Model of Meaning}
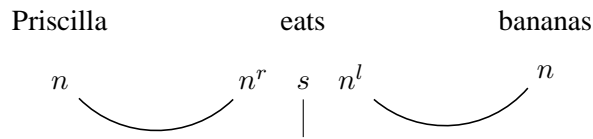
\begin{figure}[h]
\centering
\begin{tikzpicture}[node distance = 0.25cm and 2cm, text height = 1.5ex, bend angle = 45]
\node(pris) {Priscilla};
\node(pris_n) [below= of pris] {$n$};
\node(eats)[right= of pris] {eats}; 
\node(eats_s)[below= of eats] {$s$};
\node(end1)[below=0.5cm of eats_s]{};
\node(n_eats)[left=0.1cm of eats_s] {$n^r$}
	edge [-, bend left, semithick] (pris_n);
\node(eats_n)[right=0.1cm of eats_s] {${n^l}$};
\node(ban)[right=of eats] {bananas}; 
\node(ban_n)[below=0.1cm of ban] {$n$}
	edge [-, bend left, semithick] (eats_n);

\draw[-] (eats_s) -- (end1);
\end{tikzpicture}
\caption{The reduction $n n^r s n^l n \leq s$ represented in the graphical calculus.}
\label{fig:string_dia}
\end{figure}

We summarise the DisCo model: for a full description see \cite{coecke2010}. The model characterizes individual words as co-occurrence vectors. Composite terms are constructed based on the grammatical structure described by a Lambek pregroup grammar. Both pregroup grammars and vector spaces have a compact closed monoidal structure, and therefore the operations in the grammar can be transferred to vector spaces. The grammar includes atomic types $n$ for nouns and $s$ for sentences, with adjoints $x^r$, $x^l$ such that: $x^l x \leq 1 \leq x x^l \text{ and } x x^r \leq 1 \leq x^r x$, and composite types such as verb = $n^r s n^l$. A string of words is represented by the juxtaposition of the relevant types. For example, ``Priscilla eats bananas'' is rendered as $n (n^r s n^l) n$. The reduction rules are applied, and if the resulting type is an $s$, the sentence is judged grammatical. In the given example, we have $n n^r s n^l n \leq 1\cdot s n^l n \leq 1 \cdot s \cdot 1 \leq s$. Reductions can be described in a graphical calculus, shown in figure \ref{fig:string_dia}. 

Within the category of vector spaces, concatenation of atomic types into composite types, and words into sentences, is rendered as a tensor product. The reduction map $\epsilon$ is an inner product.

\section{Integrated Connectionist/Symbolic Cognitive Architecture}
In \cite{smolensky2006}, symbols are represented in a high dimensional vector space as the tensor product of filler vectors $\mathbf{f}_i$, representing meaning, and role vectors $r_i$, which specify type. These composite filler/role vectors are concatenated by summation. Filler/role vectors may also be recursively embedded, and structures such as sentences may be represented as a sum $\mathbf{s} = \sum_i{\mathbf{f}_i \otimes r_i}$.

Grammars are represented by matrices $\mathbb{W} = \sum_i w_i \mathbb{W}_i$. Rules are encoded in each $\mathbb{W}_i$ so that the calculation $H(\mathbf{s}) = \mathbf{s}^\top\mathbb{W}\mathbf{s}$ returns a weighted sum of the number of times each rule occurs in the sentence. The quantity $H(\mathbf{s})$ is called \emph{harmony}, and provides a graded measure of well-formedness of a sentence. For example, a harmonic grammar for intransitive sentences could have the rules 1) $S \rightarrow N~V$ and 2) ``$S$ is at the root", encoded in weight matrices $\mathbb{W}_{S\rightarrow N V}$ and $\mathbb{W}_S$. The sentence $\mathbf{s} = $ ``John runs" obeys both these rules, so if each rule has weighting 1, we assign the sentence a harmony value $H(\mathbf{s}) = 2$. However, the sentence ``runs John'' does not obey the rule $S \rightarrow N~V$, and so we assign it harmony $H(\mathbf{t}) = 1$. We can therefore judge ``John runs'' to be more grammatical than ``runs John''. These calculations can be performed using a vector representation $\mathbf{s}$ of the sentence and matrix representation $(\mathbb{W}_S + \mathbb{W}_{S \rightarrow N V})$ of the grammar, so that $H(\mathbf{s}) = \mathbf{s}^\top (\mathbb{W}_S + \mathbb{W}_{S \rightarrow N V}) \mathbf{s}$.

\section{Comparisons with the DisCo Model}
S\&L introduce a graded notion of grammaticality, whereas grammaticality in pregroup grammar is binary. However, we can draw parallels between the use of the rule matrices in the ICS architecture and the operations of the DisCo model. In S\&L's framework, components of a sentence interact indirectly through the rules of the grammar. In contrast, components of sentences in the DisCo model interact directly. Further, rule matrices in the ICS architecture have a similar structure to components of a sentence. We could consider the rule matrices themselves to be components of the sentence, and matrix multiplication to act as cancellation between components.

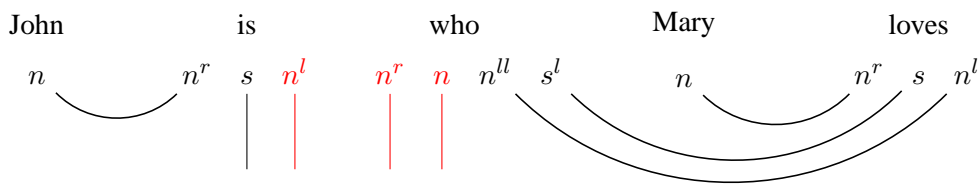
\begin{figure}[h]
\centering
\begin{tikzpicture}[node distance = 0.125cm and 2cm, text height = 1.5ex, bend angle = 45]
\node(john) {John};
\node(john_n) [below= of john] {$n$};
\node(is)[right= of john] {is}; 
\node(is_s)[below= of is] {$s$};
\node(end1)[below=1cm of is_s]{};
\node(n_is)[left=0.1cm of is_s] {$n^r$}
	edge [-, bend left, semithick] (john_n);
\node(is_n)[right=0.1cm of is_s] {$\textcolor{red}{n^l}$};
\node(end2)[below=1cm of is_n]{};
\node(who)[right=of is] {who}; 
\node(nr_who)[below left= 0.125cm and 0.05cm of who] {$\textcolor{red}{{n^r}}$};
\node(end3)[below=1cm of nr_who]{};
\node(n_who)[right=0.1cm of nr_who] {$\textcolor{red}{n}$};
\node(end4)[below=1cm of n_who]{};
\node(who_nll)[right=0.1cm of n_who] {$n^{ll}$};
\node(who_sl)[right=0.1cm of who_nll] {$s^l$};
\node(mary) [right = of who]{Mary};
\node(mary_n) [below= of mary] {$n$};
\node(loves)[right= of mary] {loves}; 
\node(loves_s)[below= of loves] {$s$}
	edge [-, bend left, semithick] (who_sl);
\node(n_loves)[left=0.1cm of loves_s] {$n^r$}
	edge [-, bend left, semithick] (mary_n);
\node(loves_n)[right=0.1cm of loves_s] {$n^l$}
	edge [-, bend left, semithick] (who_nll);

\draw[-] (is_s) -- (end1);
\draw[-, color=red] (is_n) -- (end2);
\draw[-, color=red] (nr_who) -- (end3);
\draw[-, color=red] (n_who) -- (end4);
\end{tikzpicture}
\caption{Reduction of an ungrammatical sentence. Red lines indicate unreduced types.}
\label{fig:ungram}
\end{figure}

We may then view rules of the form $\mathbb{W}_{X \rightarrow AB}$ as corresponding to one or more $\epsilon$ reductions, and $\mathbb{W}_S$ to the fact that the sentence should reduce to $s$. We therefore define a simple harmonic grammar in the DisCo framework by assigning one point to an $\epsilon$-reduction and one to reducing to the type $s$. Then the ungrammatical sentence ``John is who Mary loves" (figure \ref{fig:ungram}),  can be given a measure of grammaticality. 

The contribution to grammaticality of other operations in the DisCo framework, such as type introduction or operations within Frobenius algebra \cite{clark2013} can be investigated. Further questions include how word sequences could be rendered grammatical, for example by adding, removing, or permuting words, and how a graded measure could aid search in production of grammatical sentences.
\bibliographystyle{plainnat}

\end{document}